\pdfoutput=1
\documentclass[11pt,a4paper]{article}
\usepackage[hyperref]{acl2020}

\usepackage{times}
\usepackage{subfig}
\usepackage{array}
\usepackage{multirow}
\usepackage{graphicx}
\usepackage{latexsym}
\usepackage{amssymb}
\usepackage{amsmath}
\usepackage{booktabs}
\usepackage{url}
\usepackage{comment}
\usepackage{lipsum}
\usepackage{enumitem}
\usepackage[T1]{fontenc}
\usepackage{hyphenat}

\setlist[itemize]{noitemsep, leftmargin=*}
\setlist[enumerate]{leftmargin=*}

\usepackage{microtype}

\aclfinalcopy 
\def\aclpaperid{1767} 

\title{
    Will-They-Won't-They: \\ A Very Large Dataset for Stance Detection on Twitter}

\author{
    Costanza Conforti$^1$,~
    Jakob Berndt$^2$,~
    Mohammad Taher Pilehvar$^{1,3}$,\\
    \textbf{Chryssi Giannitsarou$^2$,~
    Flavio Toxvaerd$^2$,~
    Nigel Collier$^1$} \\[3pt]
    $^1$ Language Technology Lab, University of Cambridge\\
    $^2$ Faculty of Economics, University of Cambridge\\
    $^3$ Tehran Institute for Advanced Studies, Iran\\[3pt]
    \texttt{\{cc918,jb2088,mp792,cg349,fmot2,nhc30\}@cam.ac.uk}}
\date{}

\begin{document}

\urlstyle{same}

\maketitle

\begin{abstract}

We present a new challenging stance detection dataset, called Will-They-Won't-They\footnote{\label{note1}\url{https://en.wiktionary.org/wiki/will-they-won\%27t-they}} (\textsc{wt--wt}), which contains 51,284 tweets in English, making it by far the largest available dataset of the type.~%
All the annotations are carried out by experts; therefore, the dataset constitutes a high-quality and reliable benchmark for future research in stance detection.~%
Our experiments with a wide range of recent state-of-the-art stance detection systems show that the dataset poses a strong challenge to existing models in this domain.
The entire dataset is released for future research\footnote{\url{https://github.com/cambridge-wtwt/acl2020-wtwt-tweets}}.

\end{abstract}

\section{Introduction}\label{sec:intro}

Apart from constituting an interesting task on its own, stance detection 
has been identified as a crucial sub-step towards many other NLP tasks \cite{DBLP:journals/toit/MohammadSK17}. 
In fact, stance detection is the core component of fake news detection~\cite{pomerleau2017fake}, fact-checking~\cite{DBLP:conf/acl/VlachosR14, DBLP:conf/naacl/BalyMGMMN18}, and  rumor verification~\cite{DBLP:journals/ipm/ZubiagaKLPLBCA18}.

Despite its importance, stance detection suffers from the lack of a large dataset which would allow for reliable comparison between models.
We aim at filling this gap by presenting Will-They-Won't-They~(\textsc{wt--wt}), a large dataset of English tweets targeted at stance detection for the rumor verification task.
We constructed the dataset based on tweets, since Twitter is a highly relevant platform for rumour verification, which is popular with the public as well as politicians and enterprises~\cite{gorrell-etal-2019-semeval}.

To make the dataset representative of a realistic scenario, we opted for a real-world application of the rumor verification task in finance.
Specifically, we constructed the dataset based on tweets that discuss mergers and acquisition~(M\&A) operations between companies.
M\&A is a general term that refers to various types of financial transactions in which the ownership of companies are transferred. 
An  M\&A  process  has  many  stages  that range  from  informal  talks  to  the  closing  of  the deal. The discussions between companies are usually not publicly disclosed during the 
early stages of the process~\cite{bruner2004applied,piesse2013merger}.
In this sense, the analysis of the evolution of opinions and concerns expressed by users about a possible M\&A deal, from its early stage to its closing (or its rejection) stage, is a process similar to rumor verification~\cite{DBLP:journals/csur/ZubiagaABLP18}.

Moreover, despite the wide interest, 
most 
research in the intersection of NLP and finance has so far focused on sentiment analysis, text mining and thesauri/taxonomy generation~\cite{DBLP:journals/isafm/FisherGH16, hahn2018proceedings, financialLREC}.
While sentiment~\cite{chan2017sentiment} and targeted-sentiment analysis~\cite{DBLP:conf/semeval/ChenHC17} have an undisputed importance for analyzing financial markets, 
research in stance detection takes on a crucial role:
in fact, 
being able to model the market's perception of 
the merger might ultimately contribute to explaining stock price re-valuation.

We make the following three contributions.
Firstly, we construct and release \textsc{wt--wt}, a large, expert-annotated Twitter stance detection dataset.
With its 51,284 tweets, the dataset is an order of magnitude larger than any other stance detection dataset of user-generated data, and could be used to train and robustly compare neural models.
To our knowledge, this is the first resource for stance in the financial domain.
Secondly, we  demonstrate the utility of the \textsc{wt--wt} dataset
by evaluating
11 competitive and state-of-the-art stance detection models on our benchmark. 
Thirdly, we annotate a further M\&A operation in the entertainment domain; we investigate the robustness of best-performing models on this operation, and show that such systems struggle
even over small domain shifts. The entire dataset is released to enable research in stance detection and domain adaptation.

\section{Building the \textsc{wt--wt} Dataset}\label{sec:data}

We consider five recent 
operations, 4 in the healthcare and 1 in the entertainment industry~
(Table~\ref{tab:corpora}). 
\begin{table}[t]
    \setlength{\tabcolsep}{3pt}
    \small
    \centering
    \scalebox{1}{
\begin{tabular}{llll}
\toprule
{\bf M\&A} & {\bf Buyer} & {\bf Target} & {\bf Outcome}\\
\midrule
\textsc{cvs\_aet} & CVS Health &Aetna & Succeeded \\
\textsc{ci\_esrx} & Cigna& Express Scripts & Succeeded \\
\textsc{antm\_ci} & Anthem & Cigna & Blocked \\
\textsc{aet\_hum}& Aetna & Humana & Blocked \\
\textsc{dis\_foxa}& Disney & 21st Century Fox & Succeeded \\
\bottomrule
\end{tabular}
}
    \caption{Considered M\&A operations. Note that \textsc{aet} and \textsc{ci} appear both as buyers and as targets.}
    \label{tab:corpora}
\end{table}

\smallskip
\noindent
\subsection{Data Retrieval} 
For each operation, we used Selenium\footnote{\url{www.seleniumhq.org}} to retrieve IDs of tweets with  
one of the following sets of keywords: mentions of both companies' names or acronyms, and mentions of one of the two companies 
with a set of merger-specific terms (refer to Appendix A.1 for further details). 
Based on historically available information about M\&As, we sampled 
messages from one year before the proposed merger's date up to six months after the merger 
took place.
Finally, we obtain the text of a tweet by crawling for its ID using Tweepy\footnote{\url{www.tweepy.org/}}.

\smallskip
\noindent

\subsection{Task Definition and Annotation Guidelines}
The annotation process was preceded by a pilot annotation, after which the final annotation guidelines were written in close collaboration with three domain experts. We followed the convention in Twitter stance detection~\cite{DBLP:journals/toit/MohammadSK17} and considered three stance labels: \textit{support}, \textit{refute} and \textit{comment}. We also added an \textit{unrelated} tag, obtaining the following label set:
\smallskip
\begin{enumerate}[noitemsep,topsep=0pt]
    \item 
    {Support}: the tweet is stating that the two companies will merge.\\
    \textsc{[ci\_esrx]} \textit{Cigna to acquire Express Scripts for \$52B in health care shakeup via usatoday}
    \item 
    {Refute}: the tweet is voicing doubts that the two companies will merge.\\
    \textsc{[aet\_hum]} \textit{Federal judge rejects Aetna's bid to buy Louisville-based Humana for \$34 billion} 
    \item
    {Comment}: the tweet is commenting on merger, neither directly supporting, nor refuting it.\\
   \textsc{[ci\_esrx]} \textit{Cigna-Express Scripts deal unlikely to benefit consumers}
    \item 
    {Unrelated}: the tweet is unrelated to merger.\\
    \textsc{[cvs\_aet]} \textit{Aetna Announces Accountable Care Agreement with Weill Cornell Physicians}
\end{enumerate}
\smallskip
The obtained four-class annotation schema is similar to those in other corpora for news stance detection~\cite{DBLP:conf/coling/HanselowskiSSCC18,DBLP:conf/naacl/BalyMGMMN18}.
Note that, depending on the given target, the same sample can receive a different stance label:
\smallskip
\begin{itemize}[noitemsep,topsep=0pt]
\item[\textcolor{white}{\textbullet}] \textit{Merger hopes for Aetna\hyp{}Humana remain, Anthem\hyp{}Cigna not so much.}\\
\textsc{[aet\_hum]} $\rightarrow$ \textit{support}\\
\textsc{[antm\_ci]} $\rightarrow$ \textit{refute}
\end{itemize}
\smallskip
\noindent
As observed in~\citet{DBLP:journals/toit/MohammadSK17}, stance detection is different but closely related to targeted sentiment analysis, which considers the emotions
conveyed in a text~\cite{ahothali_good_2015}.
To highlight this subtle difference, consider the following sample:
\smallskip
\begin{itemize}[noitemsep,topsep=0pt]
\item[\textcolor{white}{\textbullet}]
\textsc{[cvs\_aet]} \textit{\#Cancer patients will suffer if @CVSHealth buys @Aetna CVS \#PBM has resulted in delfays in therapy, switches, etc 
-- all documented. 
Terrible!}
\end{itemize}

\smallskip
\noindent
While its sentiment towards the target operation is  \textit{negative} (the user believes that the merger will be harmful for patients), following the guidelines, its stance should be labeled as \textit{comment}: the user is talking about the implications of the operation, without expressing the orientation that the merger will happen (or not).
Refer to Appendix A.2 for a detailed description of the four considered labels.

\begin{table*}[ht]
    \centering\small
    \setlength{\tabcolsep}{5pt}
\begin{tabular}{lcccccccccc}\toprule
          \multirow{4}{*}{Label} & 
          \multicolumn{8}{c}{\bf Healthcare} &
          \multicolumn{2}{c}{\bf Entertainment} \\
          
            \cmidrule(l{2pt}r{5pt}){2-9}
            \cmidrule(l{5pt}r{2pt}){10-11}

          &
          \multicolumn{2}{c}{\textsc{CVS\_AET}} &
          \multicolumn{2}{c}{\textsc{CI\_ESRX}} &
          \multicolumn{2}{c}{\textsc{ANTM\_CI}} &
          \multicolumn{2}{c}{\textsc{AET\_HUM}} &  
          \multicolumn{2}{c}{\textsc{DIS\_FOXA}}
          \\
            \cmidrule(l{2pt}r{2pt}){2-3}
            \cmidrule(l{2pt}r{2pt}){4-5}
            \cmidrule(l{2pt}r{2pt}){6-7}
            \cmidrule(l{2pt}r{5pt}){8-9}
            \cmidrule(l{5pt}r{2pt}){10-11}
        & \# samples & \% & \# samples & \% & \# samples & \% & \# samples & \% & \# samples & \%  \\
    \midrule
    support &  
    2,469  &  21.24 &
    773  &  30.58 &
    {\color{white}0}970  &  {\color{white}0}8.78 &
    1,038  &  13.14 &
    {\color{white}0}1,413  & {\color{white}0}7.76\\
    refute &
        ~~518  &  {\color{white}0}4.45 &
        253  &  10.01 &
        1,969  &  17.82 &
        1,106  &  14.00 &
        {\color{white}0} ~~378    & {\color{white}0}2.07\\
    comment & 
        5,520  &  47.49 &
        947  &   37.47 &
        3,098  &  28.05 &
        2,804  &  35.50 &
        {\color{white}0} 8,495  & 46.69\\
    unrelated & 
        3,115  &  26.80 &
        554  &   21.92 &
        5,007  &  45.33 &
        2,949  &  37.34 &
        {\color{white}0} 7,908  & 43.46\\
    \midrule
    total & 
        11,622  && 
        {\color{white}0}2,527   && 
        11,622  && 
        {\color{white}0}7,897   && 
        18,194\\
\bottomrule
      \end{tabular}
      \caption{Label distribution across different M\&A operations (Table \ref{tab:corpora}): 
      four mergers in the healthcare domain (33,090 tweets) and one merger in the entertainment domain. The total number of 
      tweets is: 51,284.}
    \label{tab:stats_distr}
\end{table*}

\begin{table}[!ht]
\setlength{\tabcolsep}{4pt}
    \centering\small
    \scalebox{0.95}{
    \begin{tabular}{lrr}
    \toprule
    \multicolumn{2}{r}{\bf total twt} & {\bf avg twt/target}
    \\
    \midrule
    \citet{DBLP:conf/semeval/MohammadKSZC16} & 
    4,870 & 811\\
    \citet{DBLP:conf/eacl/InkpenZS17} 
    & 4,455 & 1,485\\
    \citet{DBLP:conf/socinfo/AkerZBKPL17} 
    & 401 & 401\\
    \citet{DBLP:conf/semeval/DerczynskiBLPHZ17}
    &5,568 & 696\\
    \citet{gorrell-etal-2019-semeval} {(only Twitter)} 
    & 6,634 & 829\\
    \midrule
     \textsc{wt--wt} 
     & 51,284 & 10,256 \\
    \bottomrule
    \end{tabular}
    }
   \caption{Statistics of Twitter stance detection datasets.
   }
    \label{tab:corpora_comparison}
\end{table}

\smallskip
\noindent
\subsection{Data Annotation}
During the annotation process, each tweet was independently labeled by 2 to 6 annotators. Ten experts in the financial domain were employed as annotators\footnote{Two MPhil, six PhD students and two lecturers at the Faculty of Economics of the University of Cambridge}.
Annotators received tweets in batches of 2,000 samples at a time, and were asked to annotate no more than one batch per week. The entire annotation process lasted 4 months.
In case of disagreement, the gold label was obtained through majority vote, discarding samples where this was not possible (0.2\% of the total).

\smallskip
\noindent
\subsection{Quality Assessment}
The  average Cohen's $\kappa$ between the annotator pairs\footnote{
The average $\kappa$ was weighted by the number of samples annotated by each pair. 
The standard deviation of the $\kappa$ scores between single annotator pairs is 0.074.
} 0.67, which is \textit{substantial}~\cite{cohen1960coefficient}. 
To estimate the quality of the obtained corpus, 
a further domain-expert 
labeled a random sample of 3,000 tweets, which were used {as human upperbound for evaluation (Table~\ref{tab:results_single_operation})}.
Cohen's $\kappa$ between those labels and the gold is 0.88. 
{
This is well above the agreement obtained in previously released datasets where crowd-sourcing was used (the agreement scores reported, in terms of percentage, range from 63.7\%~\cite{DBLP:conf/semeval/DerczynskiBLPHZ17} to 79.7\%~\cite{DBLP:conf/eacl/InkpenZS17}).}

{
\textit{Support-comment} samples constitute the most common source of disagreement between annotators:
this might indicate that such samples are the most subjective to discriminate, and might also contribute to explain 
the high number of misclassifications between those classes which have been observed in other research efforts on stance detection~\cite{DBLP:conf/coling/HanselowskiSSCC18}.
Moreover, w.r.t.~stance datasets where unrelated samples were randomly generated~\cite{pomerleau2017fake,DBLP:conf/coling/HanselowskiSSCC18}, we report a slightly higher disagreement between \textit{unrelated} and \textit{comment} samples, indicating that our task setting is more challenging.}

\begin{table*}[ht]
    \centering\small
    \setlength{\tabcolsep}{6.5pt}
    \scalebox{0.97}{
    \begin{tabular}{p{1.1cm}
    >{\centering\arraybackslash}m{1.4cm}
    >{\centering\arraybackslash}m{1.4cm}
    >{\centering\arraybackslash}m{1.4cm}
    >{\centering\arraybackslash}m{1.4cm}
    >{\centering\arraybackslash}m{0.8cm}
    >{\centering\arraybackslash}m{0.8cm}
    >{\centering\arraybackslash}m{0.7cm}
    >{\centering\arraybackslash}m{0.7cm}
    >{\centering\arraybackslash}m{0.7cm}
    >{\centering\arraybackslash}m{0.7cm}
    }\toprule
    
    & \multicolumn{4}{c}{\bf Macro $F_1$ across healthcare opertations}
    & & &
    \multicolumn{4}{c}{\bf Average per-class accuracy} \\
    
    \cmidrule(l{2pt}r{2pt}){2-5}
    \cmidrule(l{8pt}r{2pt}){8-11}
    
    Encoder %
    &\textsc{cvs\_aet}&
    \textsc{ci\_esrx}& 
    \textsc{antm\_ci}&
    \textsc{aet\_hum} & 
    $avgF_1$ & $avg_{w}F_1$ & 
    \textit{sup} & 
    \textit{ref} &
    \textit{com} &
    \textit{unr}\\
    \cmidrule(l{2pt}r{2pt}){1-1}
    \cmidrule(l{2pt}r{2pt}){2-5}
    \cmidrule(l{2pt}r{2pt}){6-7}
    \cmidrule(l{8pt}r{2pt}){8-11}
    SVM & 
51.0&51.0&65.7&65.0 &
58.1  &  58.5  & 
54.5  %
& 43.9  %
& 41.2  %
&
\textbf{88.4}  %
    \\

    MLP & 
    46.5&46.6&57.6&59.7 &
    52.6  & 52.7  & 
     55.7  %
     & 40.3  %
     & 48.6  %
     & 68.1  %
    \\

    EmbAvg & 
  50.4&51.9&50.4&58.9 &
  52.9  &  52.3  & 
 55.2  %
 & 50.5  %
 & 52.7  %
 & 67.4  %
    \\

CharCNN &
49.6&48.3&65.6&60.9 &
56.1  &  56.8  
& 55.5  %
& 44.2  %
& 41.6  %
& 82.1  %
\\
WordCNN &
46.3&39.5&56.8&59.4 
&  50.5  &  51.7  &  
62.9  
& 37.0  %
& 31.0  %
& 71.7  %
\\
BiCE & 
56.5&52.5&64.9&63.0 
& %
59.2  &  60.1  &  
61.0  
& 48.7  %
& 45.1  %
& 79.9  %
\\
CrossNet   & 
\textbf{59.1}&54.5&65.1&62.3 
& %
60.2  &  61.1   
& 63.8  %
& 48.9  %
& 50.5  %
& 75.8  %
\\
SiamNet & 
58.3&54.4&\textbf{68.7}&\textbf{67.7}
&  \textbf{62.2}  & \textbf{63.1}
& 67.0  %
& 48.0  %
& 52.5  %
& 78.3  %
\\

CoMatchAtt   & 
54.7&43.8&50.8&50.6 
& %
49.9  &  51.6  
& \textbf{71.9}  %
& 24.4  %
& 33.7  %
& 65.9  %
\\

TAN   &
56.0&55.9&66.2&66.7 
& %
  61.2  & 61.3  
& 66.1  %
& 49.0  %
& 51.7  %
& 74.1  %
\\

HAN  &
56.4&\textbf{57.3}&66.0&67.3 
&  61.7  &  61.7 
& 67.6  
& \textbf{52.0}  %
& \textbf{55.2}  %
& 69.1  %
\\
 \cmidrule(l{2pt}r{2pt}){1-5}
 \cmidrule(l{2pt}r{2pt}){8-11}
\textit{mean} 
&
53.1 &
50.5 &
61.6 &
62.0 &
$-$ & $-$ &
61.9 &
44.2 &
45.8 &
74.6 \\ 
\midrule
\it upperbound &
75.3 &
71.2 &
74.4 &
73.7 &
74.7 &
75.2 &
80.5 & 
89.6 &
71.8 &
84.0 \\

\bottomrule
    \end{tabular}
    }
    \caption{Results on the healthcare operations in the \textsc{wt--wt} dataset.
    Macro $F_1$ scores are obtained by testing on the target operation while training on the other three.     
    $avgF_1$ and $avg_wF_1$ are, respectively, the unweighted and weighted (by operations size) average of all operations. 
   }
    \label{tab:results_single_operation}
\end{table*}

\smallskip
\noindent
\subsection{Label Distribution}
The distribution of obtained labels for each operation is reported in Table~\ref{tab:stats_distr}.
Differences in label distribution between events are usual, and have been observed in other stance corpora~\cite{DBLP:conf/lrec/MohammadKSZC16,DBLP:conf/coling/KochkinaLZ18}.
For most operations, there is a clear correlation between the relative proportion of \textit{refuting} and \textit{supporting} samples and the merger being approved or blocked by the US Department of Justice.
\textit{Commenting} tweets are more frequent than \textit{supporting} over all operations: 
this is 
in line with previous findings in financial microblogging~\cite{ŽNIDARŠIČ18.5}. 

\smallskip

\noindent
\subsection{Comparison with Existing Corpora}
The first dataset for Twitter stance detection collected 4,870 tweets on 6~political events~\cite{DBLP:conf/lrec/MohammadKSZC16} and was later used in SemEval-2016
~\cite{DBLP:conf/semeval/MohammadKSZC16}. Using the same annotation schema, \citet{DBLP:conf/eacl/InkpenZS17} released a corpus 
on the 2016 US election annotated for multi-target stance.
In the scope of \textsc{Pheme}, 
a large project on rumor resolution~\cite{DBLP:conf/um/DerczynskiB14},  \citet{DBLP:journals/corr/ZubiagaHLPT15} 
stance-annotated 325 conversational trees discussing 9 breaking news events. 
The dataset was used in RumourEval 2017~\cite{DBLP:conf/semeval/DerczynskiBLPHZ17} and was later extended with 1,066 tweets for RumourEval 2019~\cite{gorrell-etal-2019-semeval}.
Following the same procedure, \citet{DBLP:conf/socinfo/AkerZBKPL17} annotated 
401 tweets on mental disorders (Table~\ref{tab:corpora_comparison}).

This makes the proposed dataset by far the largest publicly available dataset for stance detection on user-generated data.
In contrast with \citet{DBLP:conf/lrec/MohammadKSZC16}, \citet{DBLP:conf/eacl/InkpenZS17}~and \textsc{Pheme}, where crowd-sourcing was used, only highly skilled domain experts were involved in the annotation process of our dataset.
{
Moreover, previous work on stance detection focused on a relatively narrow range of mainly political topics: in this work, we widen the spectrum of considered domains in the stance detection research with a new financial dataset.}

For these reasons, the \textsc{wt--wt} dataset
constitutes a high 
quality and robust benchmark for the research community to train and compare performance of models and their scalability, {
as well as for research on domain adaptation}. Its large size also allows for pre-trainining of models, before moving to domain with data-scarcity.

\section{Experiments and Results}\label{sec:experiments}

We re-implement 11 architectures recently proposed for stance detection. Each system takes as input a tweet and the related target, represented as a string with the two considered companies.
A detailed description of the models, with references to the original papers, can be found in Appendix B.1.
Each architecture produces a single vector representation $h$ for each input sample. Given $h$, we predict $\hat{y}$ with a softmax operation over the 4 considered labels.

\subsection{Experimental Setup}\label{subsec:setup}
We perform common preprocessing steps, such as URL and username normalization (see Appendix B.2).
All hyper-parameters are listed in Appendix B.1 for replication.
In order to allow for a fair comparison between models, they are all initialized with Glove embeddings pretrained on Twitter\footnote{\url{https://nlp.stanford.edu/projects/}}
~\cite{pennington2014glove}, 
which are shared between tweets and targets and kept fixed during training.

\subsection{Results and Discussion}
\label{sec:experiments_ subsec}
Results of experiments are reported in Table~\ref{tab:results_single_operation}.
Despite its simple architecture, SiamNet obtains the best performance in terms of both averaged
and weighted averaged 
$F_1$ scores. In line with previous findings~\cite{DBLP:journals/toit/MohammadSK17}, the SVM model constitutes a very strong and robust baseline. The relative gains in performance of CrossNet w.r.t.~BiCE, and of HAN w.r.t.~TAN, consistently 
reflect results obtained by such models on the SemEval 2016-Task 6 corpus~\cite{DBLP:conf/acl/XuPNS18,DBLP:conf/coling/SunWZZ18}. 

Moving to single labels classification, analysis of the confusion matrices shows a relevant number of 
misclassifications between the \textit{support} and \textit{comment} classes.
Those classes have been found difficult to discriminate in other datasets as well~\cite{DBLP:conf/coling/HanselowskiSSCC18}. The presence of linguistic features, as in the HAN model, may help in spotting the nuances in the tweet's argumentative structure which allow for its correct classification. This may hold true also for the \textit{refute} class, the least common and most difficult to discriminate. %
\textit{Unrelated} samples in~\textsc{wt--wt} could be about the involved companies, but not about their merger: this makes classification more challenging than in datasets containing randomly generated \textit{unrelated} samples~\cite{pomerleau2017fake}.
SVM and CharCNN obtain the best performance on \textit{unrelated} samples: this suggests the importance of character-level information, which could be better integrated into
future architectures.

Concerning single operations, \textsc{cvs\_aet} and \textsc{ci\_esrx} have the lowest average performance across models. This is consistent with higher disagreement among annotators for the two mergers.

\subsection{Robustness over Domain Shifts}

We investigate the robustness of SiamNet, the best model in our first set of experiments, and BiCE, which constitutes 
a simpler neural baseline (Section~\ref{sec:experiments_ subsec}), over domain shifts with a cross-domain experiment on an M\&A event in the entertainment business.

\smallskip
\noindent
\textbf{Data.}~We collected data for the Disney-Fox (\textsc{dis\_foxa}) merger and  annotated them 
with
the same procedure 
as in Section~\ref{sec:data}, resulting in a total of 18,428 tweets.
The obtained distribution is highly skewed towards the \textit{unrelated} and \textit{comment} class 
(Table~\ref{tab:stats_distr}). This could be due to the fact that
users are more prone to digress and joke 
when talking about the companies behind their favorite shows than when considering
their health insurance providers (see Appendix A.2).

\begin{table}[ht]
\small
    \centering
    \begin{tabular}{lcccc}
    \toprule
    \multirow{3}{*}{\bf train $\rightarrow$ test} 
    & \multicolumn{2}{c}{\bf BiCE}
    & \multicolumn{2}{c}{\bf SiamNet}\\
    \cmidrule(l{2pt}r{2pt}){2-3}
    \cmidrule(l{2pt}r{2pt}){4-5}
    & $acc$ & $F_1$ & $acc$ & $F_1$\\
    \midrule
    
    health $\rightarrow$ health & 77.69 & 76.08 & 78.51 & 77.38 \\
    health $\rightarrow$ ent &    57.32 & 37.77 & 59.85 & 40.18 \\
    \midrule
    ent $\rightarrow$ ent &       84.28 & 74.82 & 85.01 & 75.42 \\
    ent $\rightarrow$ health &    46.45 & 33.62 & 48.99 & 35.25 \\
    
    \bottomrule
    \end{tabular}
    \caption{Domain generalization experiments across entertainment (ent) and healthcare datasets. Note that the data partitions used are different than in Table~\ref{tab:results_single_operation}.}
    \label{tab:domain_adapt_results}
\end{table}

\smallskip
\noindent
\textbf{Results.}~
We train on all healthcare operations and test on \textsc{dis\_foxa} (and the contrary), considering a 70-15-15 split between train, development and test sets for both sub-domains. Results show SiamNet consistently outperforming BiCE. 
The consistent drop in performance according to both accuracy and macro-avg $F_1$ score, which is observed in all classes but particularly evident for \textit{commenting} samples, 
indicates 
strong domain dependency and room for future research.

\section{Conclusions}
We presented \textsc{wt--wt}, a large expert-annotated dataset for stance detection with over 50K labeled tweets. 
Our experiments with 11 strong models 
indicated a consistent ($>$10\%) performance gap between the state-of-the-art and human upperbound, which proves that \textsc{wt--wt} constitutes a strong challenge for current models.
{Future research directions might explore 
the usage of transformer-based models, as well as of models which exploit not only linguistic but also network features, which have been proven to work well for existing stance detection datasets~\cite{DBLP:journals/pacmhci/AldayelM19}.}

Also, the multi-domain nature of the dataset enables future research in cross-target and cross-domain adaptation, a clear weak point of current models according to our evaluations.

\section*{Acknowledgments}
We thank the anonymous reviewers of this paper for their efforts and for the
constructive comments and suggestions.
We gratefully acknowledge funding from the Keynes Fund, University of
Cambridge (grant no.~JHOQ).
CC is grateful to NERC DREAM CDT (grant no.~1945246) for partially
funding this work.
CG and FT are thankful to the Cambridge Endowment for Research in Finance (CERF).

\bibliography{acl2019}
\bibliographystyle{acl_natbib}

\smallskip

\section*{Appendix A: Dataset-related Specifications}

\subsection*{A.1 \quad Crawling Specifications}

\begin{itemize}
\setlength\itemsep{3pt}
    \item M\&A\hyp{}specific terms used for crawling: one of \texttt{merge}, \texttt{acquisition}, \texttt{agreement}, \texttt{acquire}, \texttt{takeover}, \texttt{buyout}, \texttt{integration} + mention of a given company/acronym.
    \item Crawl start and end dates: \\
    \vspace{-7pt} \\
    \begin{tabular}{lc}
    \textsc{cvs\_aet}   & 15/02/2017 $\rightarrow$ {17/12/2018}\\
    \textsc{ci\_esrx}   & 27/05/2017 $\rightarrow$ {17/09/2018}\\
    \textsc{antm\_ci}   & 01/04/2014 $\rightarrow$ {28/04/2017}\\
    \textsc{aet\_hum}   & 01/09/2014 $\rightarrow$ {23/01/2017}\\
    \textsc{dis\_foxa}  & 09/07/2017 $\rightarrow$ {18/04/2018}
    \end{tabular}
\end{itemize}

\subsection*{A.2 \quad Description and Examples of the Considered Labels}
This is an extract from the annotation guidelines sent to the annotators. 

{\hfill}

\noindent
The annotation process consists of choosing one of four possible labels, given a tweet and an M\&A operation.
The four labels to choose from are \textit{Support}, \textit{Comment}, 
\textit{Refute}, and \textit{Unrelated}.

\smallskip
\noindent
\textbf{Label 1: Support --} 
If the tweet is supporting the theory that the merger is happening. 
Supporting tweets can be, for example, one of the following:
\begin{enumerate}[noitemsep,topsep=0pt]
    \item Explicitly stating that the deal is happening:\\
    $\rightarrow$ [\textsc{ci\_esrx}] \textit{Cigna to acquire Express Scripts for \$52B in health care shakeup via \@usatoday}
    \item Stating that the deal is likely to happen:\\
    $\rightarrow$ [\textsc{cvs\_aet}] \textit{CVS near deal to buy Aetna (Via Boston Herald) \texttt{<URL>}}
    \item Stating that the deal has been cleared:\\
    $\rightarrow$ [\textsc{cvs\_aet}] \textit{\#Breaking DOJ clears \#CVS \$69Billion deal for \#Aetna.}
\end{enumerate}
\smallskip
\noindent
\textbf{Label 2: Comment --}
If the tweet is commenting on the merger. The tweet should neither directly state that the deal is happening, nor refute this. Tweets that state the merger as a fact and then talk about, e.g. implications or consequences of the merger, should also be labelled as commenting.
Commenting tweets can be, for example, one of the following:

\begin{enumerate}[noitemsep,topsep=0pt]
    \item Talking about implications of the deal:\\
    $\rightarrow$ [\textsc{ci\_esrx}] \textit{Cigna-Express Scripts deal unlikely to benefit consumers}
    \item Stating merger as fact and commenting on something related to the deal:\\
    $\rightarrow$ [\textsc{cvs\_aet}] \textit{\#biotechnology Looking at the CVSAetna Deal One Academic Sees Major Disruptive Potential}
    \item Talking about changes in one or both of the companies involved:\\
    $\rightarrow$ [\textsc{cvs\_aet}] \textit{Great article about the impact of Epic within the CVS and Aetna Merge \texttt{<URL>}}
\end{enumerate}
\smallskip
\noindent
\textbf{Label 3: Refute --}
This label should be chosen if the tweet is refuting that the merger is happening. 
Any tweet that voices doubts or mentions potential roadblocks should be labelled as refuting.
Refuting tweets can be, for example, one of the following:

\begin{enumerate}[noitemsep,topsep=0pt]
    \item Explicitly voicing doubts about the merger:\\
     $\rightarrow$ [\textsc{antm\_ci}] \textit{business: JUST IN: Cigna terminates merger agreement with Anthem}
    \item Questioning that the companies want to move forward:\\
    $\rightarrow$ [\textsc{ci\_esrx}] \textit{Why would \$ESRX want a deal with \$CI?}
    \item Talking about potential roadblocks for the merger:\\
    $\rightarrow$ [\textsc{ci\_esrx}] \textit{Why DOJ must block the Cigna-Express Scripts merger \texttt{<URL>}}
\end{enumerate}
\smallskip
\noindent
\textbf{Label 4: Unrelated --}
If the tweet is unrelated to the given merger. 
Unrelated tweets can be, for example, one of the following:

\begin{enumerate}[noitemsep,topsep=0pt]
    \item Talking about something unrelated to the companies involved in the merger:\\
    $\rightarrow$ [\textsc{dis\_foxa}] \textit{I'm watching the Disney version of Robin Hood someone tell me how I have a crush on a cartoon fox}
    \item Talking about the companies involved in the merger, however not about the merger:\\
    $\rightarrow$ [\textsc{cvs\_aet}] \textit{CVS and Aetna's combined revenue in 2016 was larger than every U.S. company's other than Wal Mart \texttt{<URL>}}
    \item Talking about a different merger:\\
    $\rightarrow$ [\textsc{cvs\_aet}] \textit{What are the odds and which one do you think it will be?  Cigna or Humana?  Aetna acquisition rumor}
\end{enumerate}


\smallskip

\section*{Appendix B: Models-related Parameters}

\subsection*{B.1 \quad Encoder's Architectures}


\begin{itemize}[noitemsep,topsep=0pt]
    \item \textbf{SVMs}: linear-kernel SVM leveraging bag~of~\textit{n}-grams (over words and characters) features.
    A similar simple system outperformed all 19 teams in the SemEval-Task 6~\cite{DBLP:journals/toit/MohammadSK17}.
    \item \textbf{MLP}: a multi-layer perceptron
    (MLP)  with one dense layer,
    taking as input the concatenation of  tweet's and target's TF-IDF representations and their cosine similarity score (similar to the model in 
    \citet{DBLP:journals/corr/RiedelASR17}).
    \item \textbf{EmbAvg}: 
    a MLP with two dense layers, taking as input the average of the tweet's and the target's word embeddings. 
    Averaging 
    embeddings 
    was proven to work well
    for Twitter data in previous papers by~\newcite{DBLP:conf/coling/ZubiagaKLPL16, DBLP:journals/corr/KochkinaLA17}, who - differently than in this paper - classified stream of tweets in a conversation tree. 
    \item \textbf{CharCNN} and \textbf{WordCNN}: two CNN models, one over character and one over words, following the work by~\citet{DBLP:conf/semeval/VijayaraghavanS16}. 

    \item \textbf{BiCE}: a similar Bidirectional Conditional Encoding model to that of \citet{DBLP:conf/emnlp/AugensteinRVB16}: the tweet is processed by a BiLSTM whose forward and backward initial states are initialized with the last states of a further BiLSTM which processed the target. 
   
   
    
    \item \textbf{CrossNet}: a BiCE
    model augmented with~self- attention and two dense layers, 
    as in the~cross-target stance
    detection
    model~\cite{DBLP:conf/acl/XuPNS18}.
    
    \item \textbf{SiamNet}: siamese networks have been recently used for fake news stance detection~\cite{DBLP:conf/comad/SantoshBS19}. Here we implement a siamese network based on a BiLSTM followed by a 
    self-attention layer~\cite{DBLP:conf/naacl/YangYDHSH16}. The obtained tweet and target vector representations are concatenated with their similarity score (following \citet{DBLP:conf/aaai/MuellerT16}, we used the inverse exponential of the Manhattan distance).

    \item \textbf{Co-MatchAtt}: we use a similar co-matching attention mechanism as in~\citet{DBLP:conf/acl/WangYJC18} to connect the tweet and the target, encoded with two separated BiLSTM layers, followed by a self-attention layer~\cite{DBLP:conf/naacl/YangYDHSH16}.
       
    \item \textbf{TAN}: a model combining a BiLSTM and a target-specific attention extractor over target-augmented embeddings~\cite{du2017stance, DBLP:conf/ecir/DeySK18}, similarly as in~\citet{du2017stance}. 
    

    \item \textbf{HAN}: we follow \citet{DBLP:conf/coling/SunWZZ18} to implement a Hierarchical Attention Network, which uses two levels of attention to leverage the tweet representation along with linguistic information (sentiment, dependency and argument).
   
\end{itemize}

\begin{table}[ht]
\begin{center}
\begin{small}
\begin{tabular}{lr}
    \toprule
    \multicolumn{2}{l}{\textit{SVM model}}\\
    Word NGrams & 1, 2, 3 \\
    Char NGrams & 2, 3, 4 \\
    
    \midrule
    \multicolumn{2}{l}{\textit{Common to all neural models}}\\
    max tweet len & 25\\
    batch size & 32\\
    max epochs & 70\\
    optimizer & \textit{Adam} \\
    Adam learning rate & 0.001\\
    word embedding size & 200 \\
    embedding dropout & 0.2\\
    \midrule
    \multicolumn{2}{l}{\textit{TFIDF--MLP model}}\\
    BOW vocabulary size & 3000\\
    dense hidden layer size & 100\\
    \multicolumn{2}{l}{\textit{EmbAvg model}}\\
    dense hidden layers size & 128\\
    \midrule
    \multicolumn{2}{l}{\textit{WordCNN model}}\\
    window size & 2, 3, 4\\
    no filters &  200\\
    dropout & 0.5\\
    \multicolumn{2}{l}{\textit{CharCNN model}}\\
    no of stacked layers & 5 \\
    window size & 7, 7, 3, 3, 3\\
    no filters & 256 \\
    dropout & 0.2\\
    \midrule
    \multicolumn{2}{l}{\textit{BiCE, CrossNet, SiamNet and TAN model}}\\
    BiLSTM hidden size & 265*2\\
    BiLSTM recurrent dropout & 0.2\\
    \midrule
    \multicolumn{2}{l}{\textit{HAN model}}\\
    max sentiment input len & 10\\
     max dependency input len & 30\\
    max argument input len& 25 \\
     BiLSTM hidden size & 128 \\
    
    \bottomrule
\end{tabular}
\caption{Hyperparameters used for training. Whenever 
reported, we used the same as in the original papers.}
\end{small}
\end{center}
\end{table}

\subsection*{B.2 \quad Preprocessing Details}
After some preliminary experiments, we found the following preprocessing steps to perform the best:
\begin{enumerate}[noitemsep, topsep=0pt]
    \item Lowercasing and tokenizing using 
    NLTK's TwitterTokenizer\footnote{\url{https://www.nltk.org/api/nltk.tokenize.html}}.
    \item Digits and URL normalization.
    \item Low-frequency users have been normalized; high frequency users have been kept, stripping the ''@`` from the token. Such users included the official Twitter accounts of the companies involved in the mergers (like~\texttt{@askanthem}), media~(\texttt{@wsj}), official accounts of US politicians (\texttt{@potus, @thejusticedept}, ...) 
    \item The \# signs have been removed from hashtags.
\end{enumerate}
\smallskip
\noindent
We keep in the vocabulary only tokens occurring at least 3 times, resulting in 19,561 entries considering both healthcare and entertainment industry. 

We use \texttt{gensim} to extract the TF--IDF vectors froms the data\footnote{\url{https://radimrehurek.com/gensim/models/tfidfmodel.html}}, which are used in the TFIDF--MLP model. 
For the HAN model, following~\citet{DBLP:conf/coling/SunWZZ18}, we use the MPQA subjective lexicon~\cite{DBLP:conf/naacl/WilsonWH05} to extract the sentiment word sequences and the Stanford Parser\footnote{\url{https://nlp.stanford.edu/software/lex-parser.html}} to extract the dependency sequences. We train an SVM model to predict argument labels on~\citet{DBLP:conf/ijcnlp/HasanN13}'s training data, and we predict the argument sentences for the \textsc{wt--wt} dataset, as discussed in~\citet{DBLP:conf/coling/SunWZZ18}.

\end{document}